# Adaptive Neuro-Fuzzy Inference System and a Multilayer Perceptron Model Trained with Grey Wolf Optimizer for Predicting Solar Diffuse Fraction

Randall Claywell [1], Laszlo Nadai [1], Felde Imre [2], Amir Mosavi [1,*]

1. Kalman Kando Faculty of Electrical Engineering, Obuda University, 1034 Budapest, Hungary; randall.claywell@rh.uni-obuda.hu; nadai@uni-obuda.hu
2. John von Neumann Faculty of Informatics, Obuda University, 1034 Budapest, Hungary; felde@uni-obuda.hu
* Correspondence: amir.mosavi@kvk.uni-obuda.hu

**Abstract:** The accurate prediction of the solar Diffuse Fraction (DF), sometimes called the Diffuse Ratio, is an important topic for solar energy research. In the present study, the current state of Diffuse Irradiance research is discussed and then three robust, Machine Learning (ML) models, are examined using a large dataset (almost 8 years) of hourly readings from Almeria, Spain. The ML models used herein, are a hybrid Adaptive Network-based Fuzzy Inference System (ANFIS), a single Multi-Layer Perceptron (MLP) and a *hybrid* Multi-Layer Perceptron-Grey Wolf Optimizer (MLP-GWO). These models were evaluated for their predictive precision, using various Solar and Diffuse Fraction (DF) irradiance data, from Spain. The results were then evaluated using two frequently used evaluation criteria, the Mean Absolute Error (MAE) and the Root Mean Square Error (RMSE). The results showed that the MLP-GWO model, followed by the ANFIS model, provided a higher performance, in both the training and the testing procedures.

## 1. Introduction

Estimation of solar irradiance is of utmost importance for the efficient operation of solar energy production [1]. Insight into the solar irradiance is beneficial to managing the solar facilities and passive energy-efficient systems [2]. The value of global irradiance consists of direct and diffuse solar irradiance and the ratio that exists between. The direct and diffuse solar irradiance are essential for estimating solar irradiance under arbitrary surface orientations [3][4], obstructed environments [5], within interior spaces [6], for building energy simulations, impact on photovoltaic systems and the photosynthesis potentials in agricultural/forestry analysis and planning [7][8]. Recent studies have shown and measured the positive effect that Diffuse irradiation has on increasing canopy light use efficiency (LUE) in the Amazon Rain Forrest and related vegetative carbon uptake [9][10][11].

Solar irradiance varies greatly with latitude, surface inclination, terrain, season and time (with different, but predictable solar positions) and is subject to unpredictable weather conditions [12]. Many models have been evaluated for their ability to predict the diffuse fraction with varying degrees of success [13]. One study statistically compared nine models, for estimation of the diffuse fraction, using 10 years (1996-2005), of hourly global and diffuse solar radiation data and only identified 3 models for further evaluation [14]. Another study considered ten models for hourly diffuse irradiation and evaluated their performance, both in their original and locally adjusted versions, against data recorded at five sites from a subtropical-temperate zone in the southern part of South America (latitudes between 30°S and 35°S). The best estimates resulted from locally adjusted multiple-predictor models, some of which can estimated the hourly diffuse fraction with uncertainty of 18% of the mean [15]. In general, most researchers agree that low solar altitudes and a low clearness index (cloudy conditions) cause problems for diffuse fraction empirical modelling and most models are site-dependent [16][17].

According to the International Energy Agency (IEA), Renewable electricity capacity is expected to grow by over 1 TW, a 46% growth, from 2018 to 2023. PV accounts for more than half of this expansion (575GW), This growth will accelerate from 2020 onward and will be driven by supportive government policies and market improvements across most regions [18]. Solar PV generates power from sunlight, transforming solar irradiance into power. The performance of PV systems is directly affected by uncertain weather conditions (cloud cover, temperature, pollution, time-of-year). This creates challenges in PV electrical generation and power output predictions.

There are existing solutions for addressing these challenges, such as, battery storage and heat storage, that can compensate for irregular PV and power production. In addition, if one could estimate how much PV power can, theoretically, be produced, within a given timescale (hourly, daily, weekly, monthly), the operational costs for Solar Power facilities could be significantly reduced. Therefore, accurate solar irradiance forecasting is critical for the efficient production of a solar related, electrical supply, in a local grid. Since PV power output is dependent on solar irradiance, solar irradiance forecasting has been a hot topic of research in the literature. Forecasting methods can be split into three basic methodologies, i.e., physical models, statistical models, machine learning (ML) models.

The prediction of a solar economy for a given location is not only important for power forecasting, but also for energy efficient buildings. These methodologies can encompass one or incorporate a combination all three of the above methodologies. Using graphical/statistical predictive methods has been around for a long time. In 1993, NREL presented a quality control, computer mapping system that illustrated the qualities of a regions solar economy and allowed for visually identifying outliers [19]. It took advantage of 2 dimensionless solar quantities called the "Diffuse Fraction" and the "Clearness Index". There has been work done involving the use of this graphic tool involving statistical methods to develop statistically superior "quality envelopes" to identify errors in solar data and map/predict a regions solar potential [20]. More recently, this graphic/statistical methodology has been fine-tuned by using the diffuse fraction, $k_d$, and the clearness, $k_t$, to provide the possibility of a new approach to solar radiation decomposition and the diffuse fraction, founded on physical-based correlations [21][22].

Physical models make predictions based on the physical characteristics that manifest themselves in weather. Statistical models are based on historical/time-series data and are more basic than the Physical models, they are often limited by assumptions based on normality, linearity and/or certain variable dependencies. Machine learning models, however, can discover and acquire the non-linear relationships between input and output data, without being explicitly programmed for the task [23][24]. The larger the volume of data and depth of the dataset offers the potential, of a very accurate Diffuse Fraction prediction. There has been preliminary work done in Southern India, involving standard pyranometers and a continuous data collection facility, to acquire solar data for three years in the city of Aligarh (27.89N, 78.08E). Their dataset was divided into two parts involving a 'Training dataset' to develop the models, while a 'Validation dataset' was used to test the models. Reasonable agreement was found between the model estimates and the measured data [25].

Three Machine Learning models will be considered in this work. Specifically, a single MLP, a hybrid ANFIS and finally, a hybrid MLP-GWO will be evaluated for prediction performance, using various irradiance data from Almeria, Spain over a period of almost 8 years. The structure of this work includes an Introduction in Section 1, Section 2 describes the Data and Methods used in this work and a detailed description of the three models and the error evaluation metrics (MAE and RMSE) used. Section 3 provides the results of the performance of the two models, an error analysis and comparison data. Section 4 is a short description of current work in the area of diffuse irradiance prediction and a discussion of the process. Finally, Section 5 presents Conclusions and future work.

## 2. Data and Methods

*2.1 Data*

The data used herein, was culled from Almería (Spain), from a horizontal rooftop located at the University of Almería (36.83N, 2.41W and 680 AMSL). Almería is in a Mediterranean Coastal Area, in the South-eastern region of Spain. This location has a high frequency of cloudless days, an average annual temperature of 17 degrees Celsius and is usually, a high humidity environment, as would be expected near the sea [26]. The Global and Diffuse irradiance data were collected via Kipp & Zonen (Model-CM11) Pyranometers. One unit had an Eppley (model SBS) Shadow-band fixed, to measure the Diffuse irradiance. The Beam Normal irradiance was measured using an Eppley Normal Incident Pyrheliometer (Model-NIP). The original data set consisted of daily sunrise to sunset hourly values centered on GMT of measured global and diffuse horizontal irradiance, and beam normal irradiance readings, were observed over a period of 2829 days (June 1, 1990 through February 28, 1998). The entire data set contained 12,435 of daylight records. The data was quality-controlled and marked for missing time-stamps, equipment/power malfunctions and other erroneous readings. The data used for input/output/validation was solar related data is found in Table 1. below, the dataset had other metrological readings, such as, Relative Humidity, etc., these were not used.

| Table 1. Inputs and Output | |
|---|---|
| Input data | Output |
| Global Irradiance (W/m$^2$) | $k_d$ |
| Beam Irradiance (W/m$^2$) | (Global/Diffuse - Diffuse Fraction) |
| Sunshine Duration Index | |
| $k_t$ (Global/Extraterrestrial - Clearance Index) | |
| k (Diffuse/Extraterrestrial) | |

*2.2. Normalization*

Normalization was performed due to the differences in the parameters range. Equation 1 presents the formula which nomalizes the parameters between -1 to +1. Accordingly, the formula employs the minimum and maximum values and produces the normalized value between -1 and +1. This process can reduce the errors arised by differences in parameters range.

$$x_N = \left[\left(\frac{x - X_{min}}{X_{max} - X_{min}}\right) \times 2\right] - 1 \tag{1}$$

where, $x_N$ is the normalized data, $X_{min}$ is the lowest number and $X_{max}$ is the highest number in the dataset.

*2.3. Methods*

*2.3.1 Multi-layered perceptron (MLP)*

MLP as a feed-forward ANN method can successfully generate the values of the output variables according to the input variables through a non-linear function. A simple architecture of an MLP model is presented by Figure 1. According to Figure 1, MLP contains three main sections. First section imports input variables, the second section is called as the hidden layer and includes set of neurons which are called as the neurons in hidden layer. The number of neurons in the hidden layer is one of the adjustable factors affecting the accuracy of the MLP model. The last layer is called as the output layer and contains the output variables [27]. Figure 1 represents the architecture of the MLP model adopted from [28].

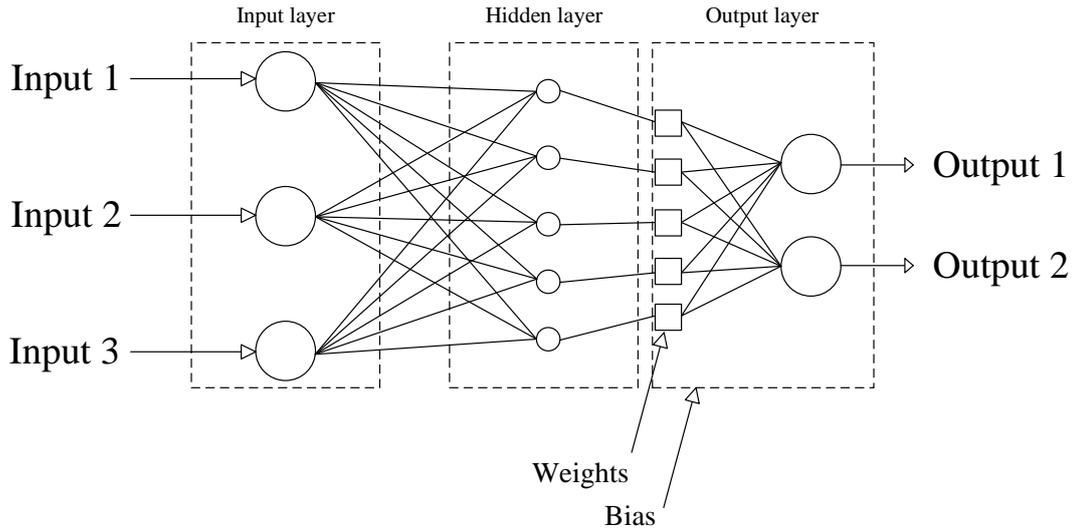

Figure 1. architecture of the MLP model

This model has been frequently described in various studies. The present section only mentions the main parts around MLP and the important points.

In a MLP a hidden layer connects the input layer to the output layer and produces the output value (f(x)) using Eq. 1, below [29]:

$$f: R^I \rightarrow R^O$$
$$f(x) = K(b^{(2)} + w^{(2)}(Q(b^{(1)} + w^{(1)}x))) \quad (1)$$

where, K and Q refer to the activation functions and b and w refer to the bias and weights, respectively. Hidden layer can be introduced by Eq. 2 [29]:

$$h(x) = Q(b^{(1)} + w^{(1)}x) \quad (2)$$

Two common activation functions for Q can be represented by Eq. 3 and 4 [29]:

$$Tanh(x) = (e^x + e^{-x})/(e^x - e^{-x}) \quad (3)$$

$$Sigmoid(x) = 1/(1 + e^{-x}) \quad (4)$$

Tanh(x) can do the task faster than the Sigmoid(x). The output vector according to [29] can be calculated by Eq. 5

$$o(x) = K(b^{(2)} + w^{(2)}h(x)) \quad (5)$$

In the present study, the architecture of the MLP has one input layer including five solar inputs:

- Global Irradiance
- Beam Normal Irradiance
- Sunshine Index
- $k_t$ (Clearance Index – Global/Extraterrestrial)
- k (Diffuse/Extraterrestrial)

There was one hidden layer including 15, 20, 25 and 30 neurons in the hidden layer, for finding the optimum number of neurons in the hidden layer and one output layer, including one output (Diffuse Fraction). The activation function was selected to be the Tanh type. Training was performed by 80% of the total data. Training was started with 15 neurons with three repetitions for finding the best run, due to the change in the results of the MLP in each training and the instability of the results in each repetition. This section wants to prepare the best architecture of the MLP to be optimized by the GWO method in the next section.

*2.3.2 MLP-GWO*

The Grey Wolf Optimizer (GWO) is known as a metaheuristic algorithm which is implemented according to the social behavior of grey wolves, while hunting. In fact, in the process of finding the best answer for the cost function is considered as the prey and hunting in the process, and wolves move towards prey with a hunting strategy. The accuracy of the algorithm depends on the population of the wolves [30]. During the hunt, grey wolves surround the prey. The following equations present the mathematical models. t refers to the current iteration, A and C refer to coefficient vectors, Xp refers to the prey position vector, and X refers to the grey wolf position vector.

$$\vec{d} = |\vec{c} \times \vec{x}_p(t) - \vec{x}(t)| \tag{6}$$

$$\vec{x}(t+1) = \vec{x}_p(t) - (2\vec{a} \times \vec{r_1} - \vec{a}) \times \vec{d} \tag{7}$$

$$\vec{c} = 2 \times \vec{r_2} \tag{8}$$

In the above relations, the variable a decreases linearly from 2 to 0 during the iterations, and r1, r2 are random vectors in the range [0, 1]. Hunting operations are usually led by Alpha. Beta and Delta wolves may occasionally hunt. In the mathematical model of Grey Wolf hunting behavior, we assumed that alpha, beta, and delta had better knowledge of the potential prey position. The first three solutions are best stored and the other agent is required to update their positions according to the position of the best search agents, as illustrated in the following equations.

$$\vec{d}_\alpha = |\vec{c_1} \times x_\alpha(t) - \vec{x}(t)|; \quad \vec{d}_\beta = |\vec{c_2} \times x_\beta(t) - \vec{x}(t)|; \quad \vec{d}_\delta = |\vec{c_3} \times x_\delta(t) - \vec{x}(t)| \tag{9}$$

$$\vec{x}(t+1) = \frac{\vec{x_1} + \vec{x_2} + \vec{x_3}}{3} \tag{10}$$

The main algorithm of the GWO can be chractrized as follows [30][31].

1. The fitness of all solutions are calculated and the top three solutions are selected as alpha, beta and delta wolves, until the algorithm is finished.

2. In each iteration, the top three solutions (alpha, beta, and delta wolves) are able to estimate the hunting position and do so in each iteration.

3. In each iteration, after determining the position of alpha, beta, delta wolves, the position of the rest of the solutions are updated by following them. During each iteration, the vectors, **a** and **c,** are updated.

4. At the end of the iterations, the position of the alpha wolf is presented as the "optimal point".

Integrateing the GWO with ANN, assures that the GWO algorithm considers the combinations of bias and weights, as the cost function, and optimizes the result ro reach the maximum efficiency [32].

*2.3.3 ANFIS*

ANFIS modellingSystem based on the comparison of the values, consists of rules, input membership functions, output membership functions, multiple inputs and an output (Fig. 4). A type of artificial neural system based on Takagi-Sugeno fuzzy interference system. Adaptive neuro fuzzy inference system (ANFIS) is used for many hybrid based data, it combines intelligent technologies to get data and produce anoutput. ANFIS is an ANN integrated by the Takagi-Sugeno fuzzy inference system. This technique was developed in the early 1990s, it has the benefits and advantages of both an ANN and a Fuzzy Inference System, it is consistent with the if-then fuzzy set of rules that can be taught to approximate nonlinear functions. Hence, ANFIS has been proposed as a universal estimator. A more detailed description of ANFIS models in terms of mathematical models are available in our recently developed studies in [33]. Figure 2 presents the main architecture of the ANFIS model which is used in the present study.

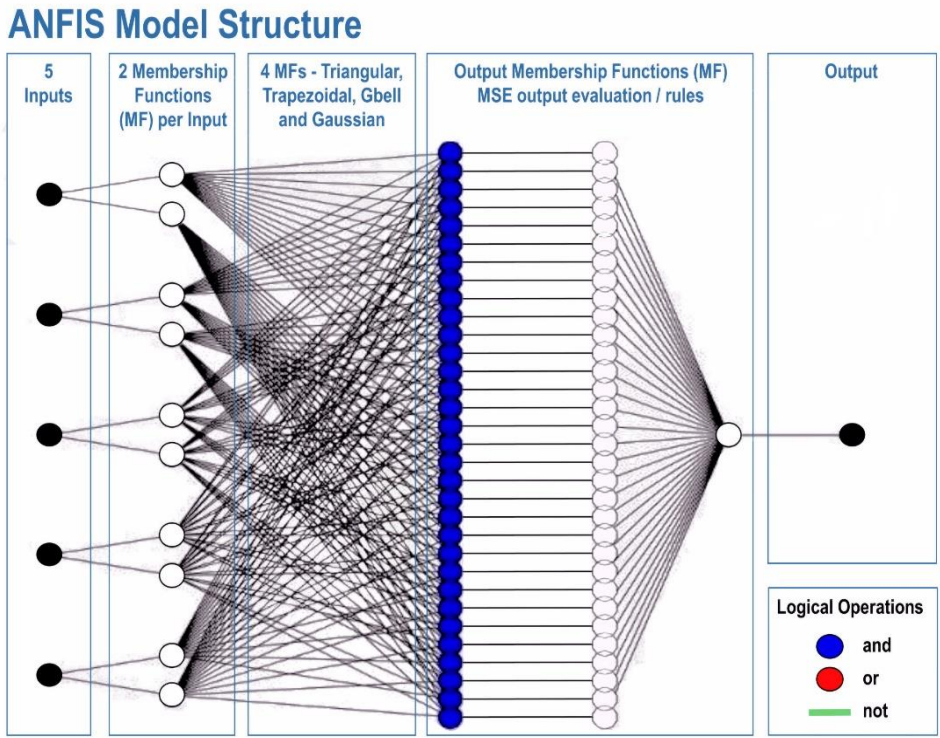

Figure 2. the architecture of the ANFIS model

The training process was initiated by five inputs using 80% of total data. Two MFs were considered for each input. Training was performed for four different types of MFs including Triangular, Trapezoidal, Gbell and gaussian MFs. In each training the output values were compared by mean square error (MSE) as the evaluation criteria factor for calculating the accuracy of the developed model. Each trainingprocess was performed during epoch number 500. The lowest MSE refers to the best prediction model. After finding the best one, the testing process was performed in the presence of the rest of the data (20%).

2.4. Evaluation criteria

The evaluation process is a step for calculating the accuracy of models, for finding the best solution for the prediction task. In the present study, the two most frequently used evaluation criteria are Mean Absolute Error (MAE) and Root Mean Square Error (RMSE). These functions employ the output and target values for calculating their distances. The following are MAE and RMSE the calculations:

$$MAE = \frac{\sum_{i=1}^{n}|x-y|}{n} \tag{12}$$

$$RMSE = \sqrt{\frac{\sum_{i=1}^{n}(x-y)^2}{n}} \tag{13}$$

Where, in both Formula 12 and 13, $x$ and $y$ are the target and predicted values, respectively and $n$, refers to the total number of data points.

## 3. Results

### 3.1 Training Results

Table 2 presents the results for MLP model. MLP was compared in the term of number of neurons in the hidden layer. As is clear, MLP with 20 neurons in the hidden layer provided the best performance compared with others. Also, 20 neurons in the hidden layer will be employed for the development of MLP-GWO.

| Table 2. MLP | | |
|---|---|---|
| No. of neurons in the hidden layer | MAE | RMSE |
| 15 | 0.329652 | 0.381277 |
| 20 | 0.283239 | 0.167089 |
| 25 | 0.303247 | 0.160102 |
| 30 | 0.294706 | 0.187014 |

Table 3 presents the results for the training phase of ANFIS model. Four main MF types including Triangular, Trapezoidal, Gbell and Gaussian MF types were employed for developing the ANFIS in training phase with two MFs and optimum method type hybrid with output MF type linear. According to the results:

| Table 3. ANFIS | | | |
|---|---|---|---|
| Description | MF type | MAE | RMSE |
| No. of MFs=2<br>Optimum method=hybrid<br>Output MF type=linear | Triangular | 0.252980 | 0.341010 |
| | Trapezoidal | 0.267428 | 0.096249 |
| | Gbell | 0.253935 | 0.089634 |
| | Gaussian | 0.251187 | 0.025520 |

MLP with 20 neurons in the hidden layer (selected from the last step), was selected to be integrated by GWO. Table 4 presents the training results for the employed MLP-GWO. The models differ in number of populations. No. of population 300, was selected as the optimum number of population with lower MAE and RMSE compared with other treatments.

| Table 4. MLP-GWO | | |
|---|---|---|
| No. of population | MAE | RMSE |
| 100 | 0.262107 | 0.343945 |
| 200 | 0.253794 | 0.093941 |
| 300 | 0.247638 | 0.088364 |
| 400 | 0.25512 | 0.097463 |

## 3.2 Testing Results

Table 5 compares the testing results of the selected models from training steps. As is clear, MLP-GWO followed by ANFIS provided the lower MAE and RMSE values.

| Table 5. Testing results | | |
|---|---|---|
| Model Name | MAE | RMSE |
| MLP | 0.503710 | 0.550427 |
| ANFIS | 0.422157 | 0.516688 |
| MLP-GWO | 0.077281 | 0.114355 |

The results are shown graphically in Figures 3,4 and 5, for each individual model and collectively, in Figure 6.

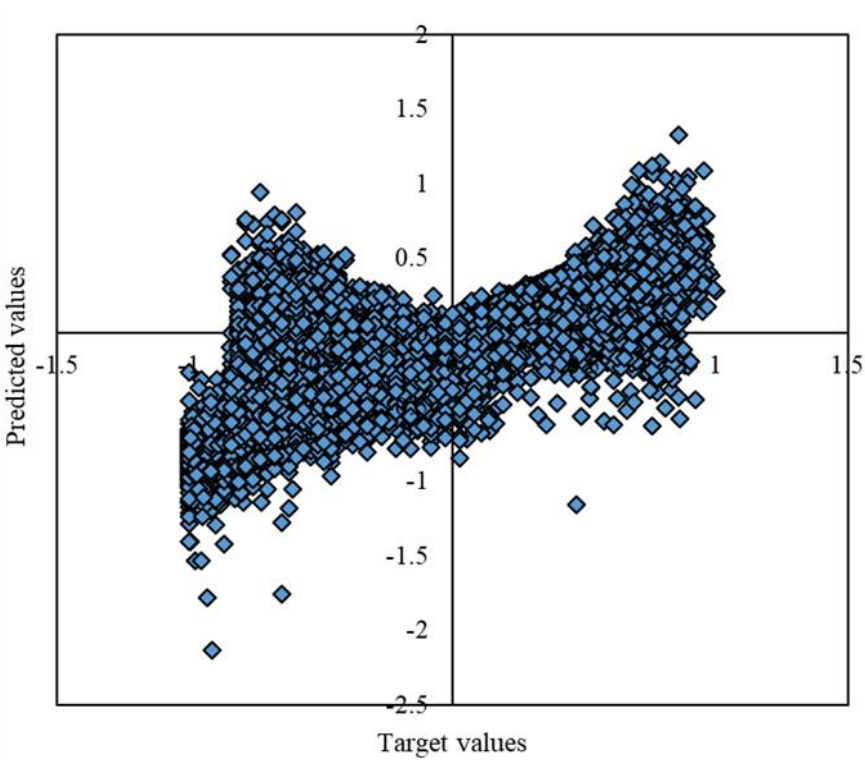

**Figure 3. Plot diagram for MLP/GWO model**

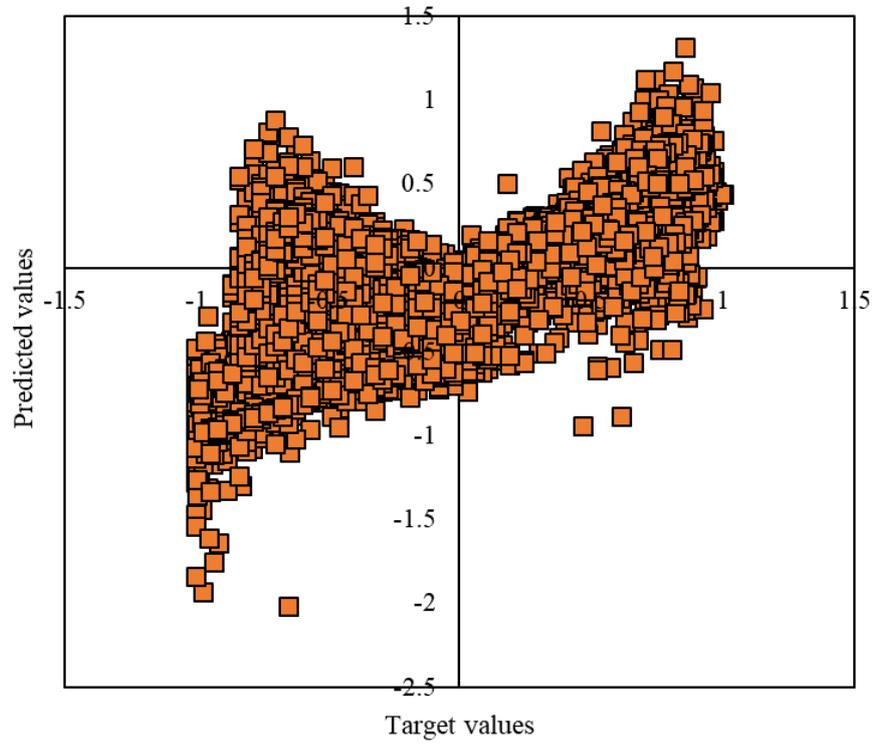

**Figure 4. Plot diagram for ANFIS model**

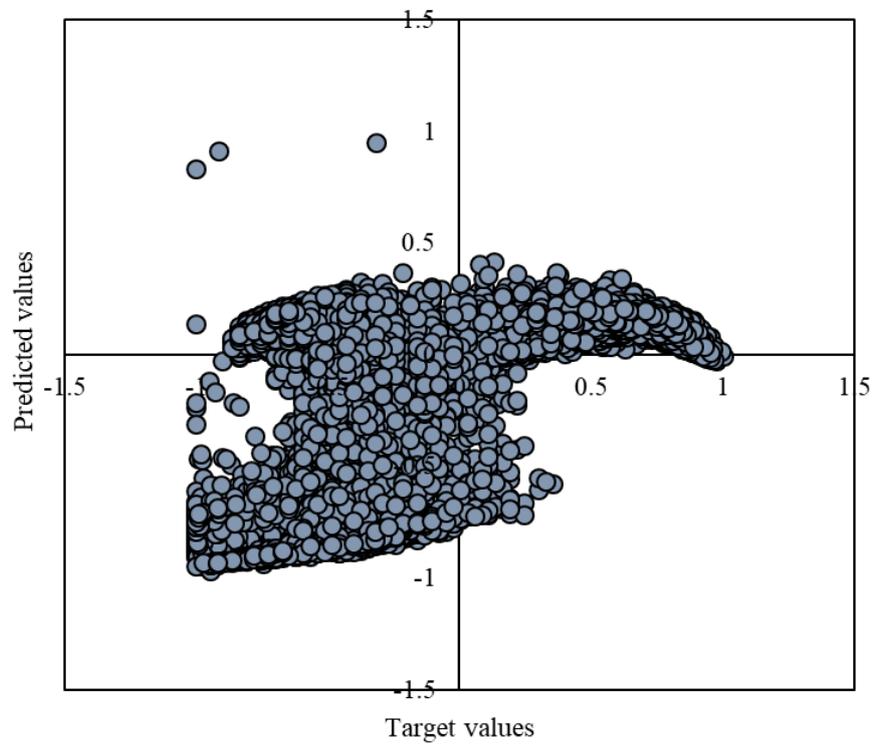

**Figure 5. Plot diagram for MLP model**

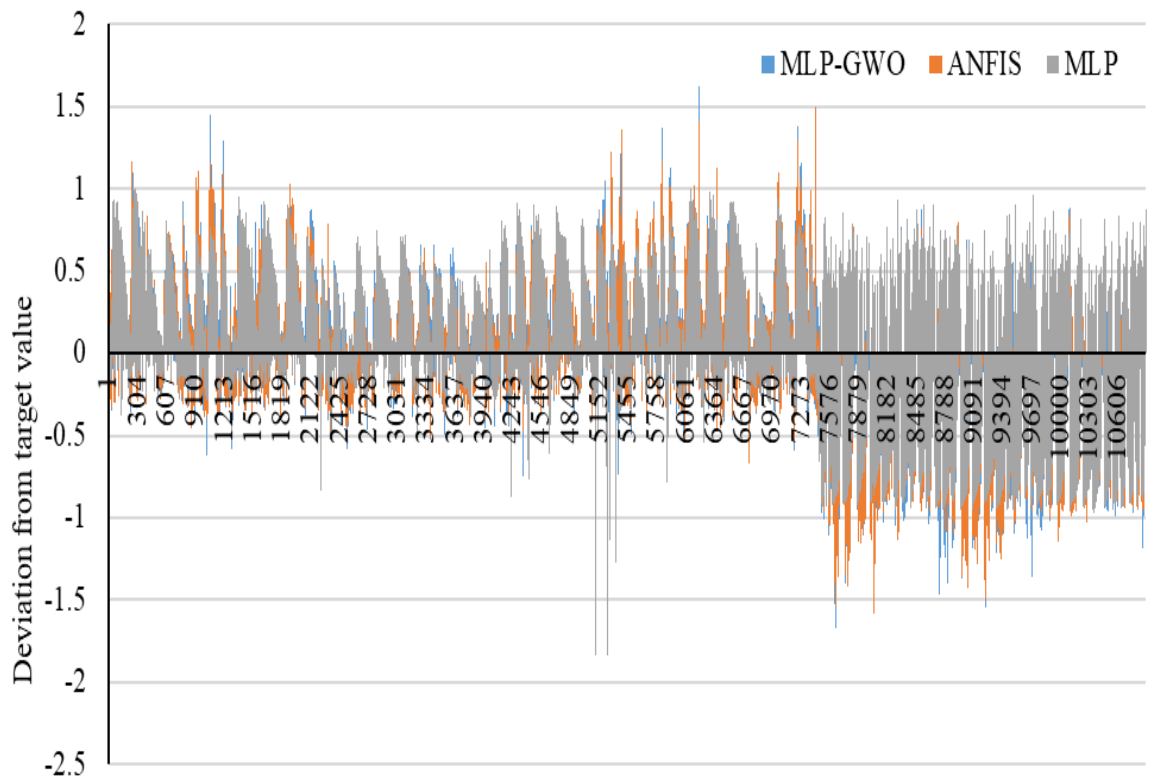

**Fig. 6. Deviations from target values for all models**

## 4. Discussion

Currently, there is important work being done in the area of Diffuse Irradiance and Diffuse Fraction data collection and prediction. This irradiance information is critical for the planning and efficient implementation of buildings, energy power systems and almost all agricultural applications. For instance, it has been shown that the Diffuse Fraction irradiance, can impact a buildings cooling by 2.3 to 5.18% in Taipei, Taiwan [34][35][36]. The accurate estimation of diffuse irradiance on a horizontal surface is emphasized by recent findings poorly calculated diffuse irradiance values can be off by as much as ±8% in over- or under-estimations for the annual energy yield of a photovoltaic systems [37]. Accurate raw data has been and is currently being remotely collected via satellite systems. The European Organization for the Exploitation of Meteorological Satellites (EUMETSAT) Satellite Application Facility for Land Surface Analysis (LSA SAF) is providing "near real-time" estimates of surface radiation data since 2005 and recent work provides diffuse fraction data, every 15 minutes for the satellite coverage areas of Europe, Africa, the Middle East, and part of South America [20][38].

In this paper, three current machine learning models are trained and evaluated for the prediction of the diffuse fraction, using recorded data, from Almeria, Spain. Diffuse fraction models are highly sensitive to local meteorological conditions and are currently, not transferable to disparate localities. The diffuse data used for this work is from an area of the world that experiences a high frequency of cloudless days and enjoys a high level solar economy, it is therefore, more predictable in nature, owing to a high clearness index. One study, from Vienna [39], evaluated 8 different diffuse fraction models and found that the top 3 models, using data from Vienna, produced a Relative Error of less than ±20%. The performance for the top 3 models was very close, showing only a slight 2% improvement, after model calibration. Using Hybrid Machine Learning and Artificial Intelligence algorithms, there seems to be room for prediction improvement in the future.

## 5. Conclusions

In the present study, three robust ML models, i.e., a MLP, ANFIS and a hybrid MLP-GWO are advanced for the prediction of the Diffuse Fraction of solar irradiance for Almeria, Spain. Results were evaluated using two frequently used evaluation criteria including MAE and RMSE. According to the results, MLP-GWO followed by ANFIS provided higher performance in both the training and the testing stages of this work. For the future research using more sophisticated hybrid machine learning models are suggested. Hybridization for training machine learning models shows significant improvement in the performance and accuracy of the models. Therefore, future models can significantly benefit from novel evolutionary algorithms and nature-inspired optimization methods to better tune the parameters of the machine leaning models.

**Author Contributions:** Conceptualization, A.M.; methodology, A.M. and S.A.; software, R.C.; validation, A.M. and S.A.; formal analysis, S.A..; investigation, R.C.; resources, A.M.; data curation, S.A.; writing—original draft preparation, A.M.; writing—review and editing, R.C.; visualization, S.A. and A.M.; supervision, N.L. and I.F.; project administration, A.M.; funding acquisition, I.F. All authors have read and agreed to the published version of the manuscript.

**Funding:** The research presented in this paper was carried out as part of the EFOP-3.6.2-16-2017-00016 project in the framework of the New Szechenyi Plan. The completion of this project ´is funded by the European Union and co-financed by the European Social Fund.

**Acknowledgments:** The research presented in this paper was carried out as part of the EFOP-3.6.2-16-2017-00016 project in the framework of the New Szechenyi Plan. The completion of this project ´is funded by the European Union and co-financed by the European Social Fund.

**Conflicts of Interest:** The authors declare no conflict of interest.